\PassOptionsToPackage{bookmarks=false}{hyperref}
\documentclass[sigconf]{acmart}

\usepackage{booktabs} % For formal tables
\usepackage{subcaption}
\graphicspath{{./figures/}}
\usepackage{amsmath}
\usepackage{amsfonts}       % blackboard math symbols
\usepackage{bbm}

\DeclareMathOperator*{\argmax}{arg\,max}
\usepackage{algorithmicx}
\usepackage{algorithm}
\usepackage{algpseudocode}

\usepackage[absolute]{textpos}

\copyrightyear{2018} 
\acmYear{2018} 
\setcopyright{acmlicensed}
\acmConference[ICCAD '18]{IEEE/ACM INTERNATIONAL CONFERENCE ON COMPUTER-AIDED DESIGN}{November 5--8, 2018}{San Diego, CA, USA}
\acmBooktitle{IEEE/ACM INTERNATIONAL CONFERENCE ON COMPUTER-AIDED DESIGN (ICCAD '18), November 5--8, 2018, San Diego, CA, USA}
\acmPrice{15.00}
\acmDOI{10.1145/3240765.3240796}
\acmISBN{978-1-4503-5950-4/18/11}

\begin{document}

\begin{textblock}{10}(4.05,1.1)
\noindent{\footnotesize \normalfont This is the authors' final version. The authoritative version will appear in the proceedings of ICCAD 2018.}
\end{textblock}

%\title{\vspace{-30pt}{\footnotesize {\normalfont This is the authors' final version. The authoritative version will appear in the proceedings of ICCAD 2018.}}\\~\\Designing Adaptive Neural Networks\\ for Energy-Constrained Image Classification}
\title{Designing Adaptive Neural Networks\\ for Energy-Constrained Image Classification}
%\titlenote{Produces the permission block, and copyright information}
%\subtitlenote{The full version \texttt{acmart.pdf} document}

\author{Dimitrios~Stamoulis, Ting-Wu (Rudy) Chin,
Anand Krishnan Prakash,\\ Haocheng Fang,
Sribhuvan Sajja, Mitchell Bognar, Diana Marculescu\\
Department of ECE, Carnegie Mellon University, Pittsburgh, PA\\
Email: dstamoul@andrew.cmu.edu }
\renewcommand{\shortauthors}{D. Stamoulis et al.}

%\IEEEoverridecommandlockouts % to allow thanks to appear
\newcommand\blfootnote[1]{%
  \begingroup
  %\footpunctfalse
  %\deffootnote{0.4cm}{0em}{\thefootnotemark~}
  \renewcommand\thefootnote{}\footnote{#1}%
  \addtocounter{footnote}{-1}%
  \endgroup
}

\begin{abstract}
As convolutional neural networks (CNNs) enable state-of-the-art computer vision applications, their high energy consumption has emerged as a key impediment to their deployment on embedded and mobile devices. Towards efficient image classification under hardware constraints, prior work has proposed adaptive CNNs, \emph{i.e.}, systems of networks with different accuracy and computation characteristics, where a selection scheme \emph{adaptively} selects the network to be evaluated for each input image. While previous efforts have investigated different network selection schemes, we find that they do not necessarily result in energy savings when deployed on mobile systems. The key limitation of existing methods is that they learn only how data should be processed among the CNNs and not the network architectures, with each network being treated as a blackbox.

To address this limitation, we pursue a more powerful design paradigm where the architecture settings of the CNNs are treated as hyper-parameters to be globally optimized. We cast the design of adaptive CNNs as a hyper-parameter optimization problem with respect to energy, accuracy, and communication constraints imposed by the mobile device. To efficiently solve this problem, we adapt Bayesian optimization to the properties of the design space, reaching near-optimal configurations in few tens of function evaluations. Our method reduces the energy consumed for image classification on a mobile device by up to $6 \times$, compared to the best previously published work that uses CNNs as blackboxes. Finally, we evaluate two image classification practices, \emph{i.e.}, classifying all images locally versus over the cloud under energy and communication constraints.
\blfootnote{$^*$ Mitchell Bognar was an intern
at Carnegie Mellon University; he is currently a student at Duke University.}
\end{abstract}

\maketitle

\section{Introduction}

Deep convolutional neural networks (CNNs) have been established
as one of the most powerful machine learning techniques for a plethora of
computer vision applications. In response to the growing demand for
state-of-the-art performance in real-world deployment, the
complexity of CNNs has increased significantly, which
has come at the significant cost of energy consumption.
As a consequence, the energy requirements of CNNs
have emerged as a key impediment
preventing their deployment on energy-constrained embedded and mobile
devices, such as Internet-of-Things (IoT) nodes, wearables, and
smartphones. For instance, object classification
with AlexNet can drain the smartphone battery
within an hour~\cite{yang2016designing}.
This design challenge has resulted in an ever-increasing interest
to develop energy-efficient image classification
solutions~\cite{ILSVRC15}.

As means to reducing the average classification
cost by trading off the accuracy, prior art has investigated
the use of adaptive CNNs, \emph{i.e.}, systems of CNNs
with different accuracy and computation characteristics,
where a selection scheme \emph{adaptively} selects the network
to be evaluated for each input image~\cite{bolukbasi17a}.
A large body of work from both industry~\cite{goodfellow2013multi} and
academia~\cite{venkataramani2015scalable, panda2016conditional,
pandaFalcon2017, parkBigLittle, panda2017energy} exploits
the insight that a large portion of an image dataset can be correctly
classified by simpler CNN architectures.
Nevertheless, all previous approaches focus only on learning
how data should be processed among the CNNs
and not the network architectures. Hence, existing
adaptive CNNs are optimized only with respect to the network selection
scheme, with each CNN being
\emph{treated as a blackbox} (\emph{i.e.}, pre-trained off-the-shelf CNN).

Moreover, prior work has not explicitly profiled energy savings
on state-of-the-art mobile systems on chip (SoCs).
Thus, the reported savings could be attributed to either significant
runtime reduction when executing smaller CNNs on powerful workstation GPUs
or to gate count (area, ergo power) reduction on ASIC-like designs. 
In our analysis, however, we found that the 
energy-limited mobile SoCs have small headroom for savings,
hence limiting the overall effectiveness of
existing adaptive CNN designs. For example, our results show that
existing methods, if tested on commercial Nvidia Tegra TX1 platforms,
could actually lead to an increase in energy consumption under strict accuracy constraints.

In this paper, we make the following observation:
the minimum energy achieved by adaptive CNNs can be
significantly reduced if the architecture of each individual network
is optimized jointly with the network selection scheme. Our \textbf{key insight} is to
treat the architectural choices of each network (\emph{e.g.},
number of hidden units, number of feature maps, \emph{etc.}) as hyper-parameters.
To the best of our knowledge, there is no global, systematic methodology for
the hardware-constrained hyper-parameter optimization of adaptive CNNs based on hardware measurements on a real mobile platform,
even though prior art has highlighted this
suboptimality~\cite{parkBigLittle}. Such observation constitutes the
key motivation behind our work.

Our work makes the following \textbf{contributions}:
%\vspace*{-5pt}
\begin{enumerate}
\item \textbf{Globally optimized adaptive CNNs}:
To the best of our knowledge, we are \emph{first} to
optimize systems of adaptive CNNs for both the network architectures
and the network selection scheme, under hardware constraints imposed by the
mobile platforms. Our methodology identifies designs that outperform
the best previously published approaches in
resource-constrained adaptive CNNs by up to $6 \times$ in terms
of minimum energy per image and by up to $31.13\%$ in terms of accuracy
improvement, when tested on a commercial Nvidia mobile SoC
and the CIFAR-10 dataset.
\item \textbf{Adaptive CNNs as hyper-parameter problem}:
To enable this design paradigm, our work formulates the design of adaptive CNNs
as a hyper-parameter optimization problem, where the architectural parameters
of each network, such as the number of hidden units and the number of feature maps,
are treated as hyper-parameters to be jointly optimized intrinsically
with respect to hardware constraints. We show that our formulation is
generic and able to incorporate different design goals that are significant
in a mobile design space. In our results, we consider optimization
under energy, accuracy, and communication constraints.

\item \textbf{Enhancing Bayesian optimization for designing adaptive CNNs}:
To solve this challenging hyper-parameter optimization problem,
we adapt Bayesian optimization (BO) to exploit properties of the mobile design space.
In particular, we observe that once the accuracy and hardware measurements (power,
runtime, energy) have been obtained for a set of candidate CNN designs, it is
relatively inexpensive to sweep over different network selection schemes to
populate the optimization process with more data points.
This \emph{fine-tuning} step allows the modified BO, which we denote
as $BO^+$ in the remainder of the paper, to reach the near-optimal
region \emph{faster} compared to a hardware-unaware BO methodology, and to effectively
reach the optimal designs identified by exhaustive (grid) search.

\item \textbf{Insightful design space exploration}:
Image classification under mobile hardware constraints
constitutes a challenging design space with several trade-offs possible.
We exploit the effectiveness of our method
as a useful aid for design space exploration, and we evaluate
two representative practices for image classification with adaptive CNNs, \emph{i.e.},
classifying all images locally or over the cloud under energy
and communication constraints. We investigate both these methods
under different energy, communication, and accuracy constraints
on a commercial Nvidia system, thus providing computer vision
practitioners and mobile platform designers with
insightful directions for future research.
\end{enumerate}

%\vspace{-3pt}
\section{Related Work}

\textbf{Adaptive CNN execution}: Prior art has shown that a large
percentage of images in an image dataset are easy to classify
even with a simpler CNN configuration~\cite{venkataramani2015scalable}.
Since the ``easy'' examples do not require the computational complexity and
overhead of a massive, monolithic CNN, this inspires the use of
adaptive CNNs with varying levels of accuracy
and complexity. This insight of dynamically trading off accuracy with energy-efficiency
can be found in several existing approaches from different groups
and under different names (\emph{e.g.}, conditional~\cite{panda2016conditional},
scalable~\cite{venkataramani2015scalable}, \emph{etc}.).

The early efforts to enable energy efficiency were based on
``early-exit'' conditions placed at each layer of a CNN,
aiming at bypassing later stages of a CNN if the
classifier has a ``confident'' prediction in earlier stages.
These methods include the scalable-effort
classifier~\cite{venkataramani2015scalable}, the conditional deep
learning classifier~\cite{panda2016conditional}, the distributed 
neural network~\cite{teerapittayanon2017distributed}, the edge-host partitioned 
neural network~\cite{ko2018edge}, and the cascading neural network~\cite{Leroux2017}.
However, recent work shows that adaptive execution at the
network level outperforms layer-level execution~\cite{bolukbasi17a}.
Hence, we focus on network-level designs. We note that
our formulation is generic and can be flexibly applied to the layer-wise case.

At the network-level, Takhirov \emph{et al.} have trained an adaptive
classifier~\cite{takhirov2016classifier}.
However, the method is applied on regression-based classifiers
of low complexity. Park \emph{et al.} propose a two-network
adaptive design~\cite{parkBigLittle}, where the decision of which
network to process the input data is done by looking into the
``confidence score'' of the network output. Bolukbasi \emph{et al.} extend the
formulation of network-level adaptive systems to multiple
networks~\cite{bolukbasi17a}. Finally, other works extend the cascading
execution to more tree-like structures for image
classification~\cite{pandaFalcon2017, roy2018tree}.

Nevertheless, existing approaches rely on off-the-shelf
network architectures without studying their architectures jointly.
While prior art has explicitly highlighted this limitation~\cite{parkBigLittle},
to the best of our knowledge there is no global, systematic methodology for
the hardware-constrained hyper-parameter optimization of adaptive CNNs.
On the contrary, our work is the first to cast this problem as a hyper-parameter optimization
problem and to enhance a BO methodology to efficiently solve it, thus
effectively designing both the CNN architectures and the chooser functions among them.

\textbf{Pruning- and quantization-based energy efficient
CNNs}: Beyond adaptive CNNs, there are two key approaches that address
energy efficiency based on CNN parameter reduction. First, prior art has
investigated pruning-based methods to reduce the network
connections~\cite{han2015learning, dai2017nest, yang2016designing}.
Second, other works reduce directly the computational complexity
of CNNs by quantizing the network weights~\cite{ding2018quantized}.
Nevertheless, recent work argues that
the effectiveness of methods targeting the weights and parameters of
a monolithic CNN relies heavily on the performance of this
baseline (seed) network~\cite{gordon2017morphnet}. Hence,
in our case we focus on adaptive CNN structures to
globally optimize and identify the sizing of the different CNNs.
Existing approaches are complimentary
and could be used to further reduce computational cost given
the globally optimal ``seed'' identified with our approach.

\textbf{Hyper-parameter optimization}:
Hyper-parameter optimization of CNNs has emerged as an increasingly challenging
and expensive process, dubbed by many researchers to be \emph{more of an art than science}.
In the context of hardware-constrained optimization,
among different model-based methods for hyper-parameter optimization,
several groups have shown that BO is effective
for architectural search in CNNs~\cite{snoek2012practical} and has been
successfully used for the co-design of hardware accelerators
and NNs~\cite{hernandez2016designing, reagen2017case},
of NNs under runtime~\cite{hernandez2016general}
and power constraints~\cite{stamoulis2017hyperpower}.
Nevertheless, BO has not been considered
in the design space of adaptive CNNs. We employ BO to
identify both the flow of input images among the CNNs
and their sizing (\emph{e.g.}, number of filters, kernel sizes, \emph{etc}.).

It is worth noting that existing methodologies rely on simplistic proxies of
the hardware performance (\emph{e.g.}, counts of the NN's weights~\cite{gelbart2014bayesian} or
energy per operation assumptions~\cite{smithson2016opal, rouhani2016delight}),
or simulation results~\cite{reagen2017case}. Since our target application
is the mobile and embedded design space, we employ BO
with all network architectures sampled and optimized for on
commercial mobile platforms (\emph{i.e.}, NVIDIA boards) based on
actual hardware measurements. In addition, our work is the
first to consider BO under constraints for both the communication
and computation energy consumption.

\section{Background}

\subsection{Adaptive Neural Networks}

Image classification in an adaptive CNN is intuitive.
To allow for inference efficiency, an adaptive CNN system
leverages the fact that many examples are correctly
classified by relatively smaller (computationally more efficient) networks,
while only a small fraction of examples require larger
(computationally expensive) networks to be correctly classified.

\begin{figure}[h!]
    \centering
    \includegraphics[width=1.0\columnwidth]{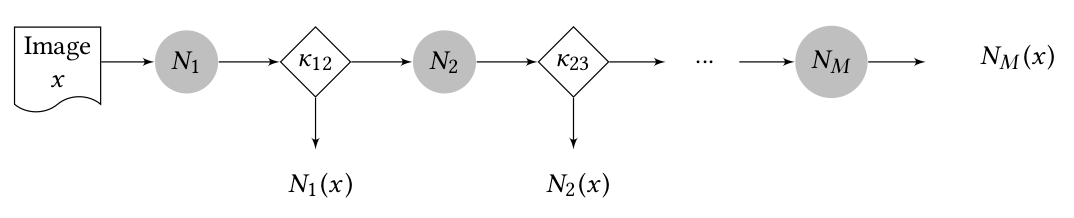}
    \vspace{-20pt}
    \caption{Classifying an image $x$ using adaptive neural networks comprised
    of $M$ CNNs.
    The system evaluates $N_i$ first, and based on a decision function $\kappa_{i,i+1}$
    decides to use $N_i(x)$ as the final prediction or to evaluate
    networks in later stages.}
    \vspace{-7pt}
  \label{fig:flow}
\end{figure}

Let us consider the case with $M$ neural networks to choose from,
\emph{i.e.} $N_1, ..., N_i, ..., N_M$, as shown in Figure~\ref{fig:flow}.
For an input data item $x \in \mathcal{X}$, we denote the
predictions of each $i$-th network as $N_i(x)~\in~R^K$, 
which represents the probabilities of input $x$ belonging to each of the $K$ classes. 
Note that $K$ is the number of classes in the classification problem.
To quantify the performance with respect to the
classification task, we denote the loss $L(\tilde{y}(x), y)$, 
where $\tilde{y}(x)$ and $y \in \{1,2,~\dots,K\}$ are the predicted and true labels, 
respectively. As common practice, $\tilde{y_i} (x)$ is defined as the most probable class
based on the current prediction, \textit{i.e.}, $\tilde{y_i}(x)=~$~argmax~$N_i(x)$.
To simplify notation, as loss we use the 0-1 loss
$L(\tilde{y}_i(x), y) = \mathbbm{1}_{\tilde{y}_i(x) \neq y}$,
showing whether the label $\tilde{y}_i(x)$ predicted from $N_i$
fails to match the true label $y$.
Let $L(\tilde{y}_i(x),y)  = L_i$ be the loss
per network $N_i$. In addition, let us denote the performance (cost)
term that we want to optimize for as $C_i = C(N_i(x))$ of the $i$-th network.

Given an image to be classified, the $N_1$ is
always executed first. Next, a decision function $\kappa$ is evaluated
to determine whether the classification decision from $N_1$ should be
returned as the final answer or the next network $N_2$ should be evaluated.
In general, we denote decision function
$\kappa_{ij}: N_i(x) \to \{0, 1\}$ that provides ``confidence feedback'' 
and decides between exiting at state
$N_i$ (\emph{i.e.}, $\kappa_{ij}=0$)
or continuing at subsequent stage $N_j$
(\emph{i.e.}, $\kappa_{ij}=1$).

\subsection{Network selection problem}

Adaptive CNNs can be seen
as an optimization problem~\cite{bolukbasi17a, parkBigLittle}.
The goal of the network selection problem is to select functions $\kappa$ that optimize for the per-image inference
cost (\emph{e.g.}, average per image runtime) subject to a constraint on the overall accuracy degradation.
For the sake of notation simplicity, we write the optimization formulation for the two-network case,
but the notation can be extended to an arbitrary number of networks $M$~\cite{bolukbasi17a}:
\begin{equation}
\label{eq:problem_generic}
\begin{split}
\underset{\theta_{12}}{\min} \text{~~}  \mathbb{E}_{x \sim \mathcal{X}} \big[ C_2
\cdot \kappa_{12} \big( N_1(x) | & \theta_{12} \big) + C_1 \big] \\
s.t. \text{~~} \mathbb{E}_{(x,y) \sim \mathcal{X}\times\mathcal{Y}} \Big[
\Big(1-\kappa_{12} \big(N_1(x) | \theta_{12} \big) \Big) \cdot & \Big( L_1 \big(\tilde{y}_1(x),y \big) - \\ 
& L_2 \big(\tilde{y}_2(x),y \big) \Big) \Big] \leq B
\end{split}
\end{equation}
where the constraint here intuitively specifies that 
whenever the image stays
at $N_1$, the error rate difference should not be greater than $B$.
We observe that in Equation~\ref{eq:problem_generic} the cost function
is optimized only over the parameters of the decision functions,
and not with respect to the parameters of the individual networks.

\textbf{Selecting the decision function $\kappa$}: An 
effective choice for the $\kappa_{ij}$ function is a
threshold-based formulation~\cite{parkBigLittle, takhirov2016classifier},
where we define as score margin $SM$ the distance between the largest and
the second largest value of the vector $N_i(x) = (N_i^1(x),
N_i^2(x), ..., N_i^K(x))$ at the output of network $N_i$.
Intuitively, the more ``confident'' the CNN is about its prediction, the larger the
value of the predicted output, thus the larger the value of $SM$
(since $N_i(x)$'s sum to 1).

Without loss of generality, in Figure~\ref{fig:flow} we show decisions between
two subsequent stages, \emph{i.e.}, $j = i+1$, for visualization and notation simplicity.
However, $N_j$ could be any other network in the design,
\emph{i.e.}, $j \in M \wedge i\neq j$, hence ``cascading'' the execution to any later stage.
For the case of $j = i+1$, let's denote $\theta_{i,i+1}$ the threshold value for the
decision function $\kappa_{i,i+1}$ between a network $N_i$ and a network $N_{i+1}$.
If $SM$ in the output of $N_i$ is larger than $\theta_{i,i+1}$, the
inference result from $N_i$ is considered correct, otherwise $N_{i+1}$ is
evaluated next, \emph{i.e.},
$\kappa_{i,i+1}(N_i(x) | \theta_{i,i+1}) = \mathbbm{1}_{[SM_i \geq \theta_{i,i+1}]}$

\subsection{Network design problem}

Existing methodologies have investigated the design of
a monolithic CNN as a hyper-parameter optimization problem~\cite{hernandez2016general,
stamoulis2017hyperpower}, where CNN architecture choices
(\emph{e.g.}, number of feature maps per convolution layer, number of neurons in fully connected layers, \emph{etc}.)
are optimally selected such that the inference cost of the CNN is minimized 
subject to maximum error constraint $b$.

To this end, we define design space $\mathcal{H}$ of a CNN,
wherein a point $ h_i = (h_{i,1}, h_{i,2}, h_{i,3}, ...) \in \mathcal{H} $
is the vector with elements the hyper-parameters of the neural network.
Intuitively, the choice of different $h$ values, \emph{i.e.},
of a different network architecture, results in 
different characteristics
in both loss $L(\tilde{y}(x|h),y)$ and computational efficiency $C(N(x|h))$
of the network. For a single network, the hyper-parameter optimization
problem is written:
\begin{equation}
\label{eq:network_design}
\begin{split}
& \underset{h \in \mathcal{H}}{\min} \text{~~} 
\mathbb{E}_{x \sim \mathcal{X}} \big[ C \big(N(x|h) \big)\big]  \text{~~~~~~~~}
s.t. \text{~~~~~~~~}  \mathbb{E}_{(x,y) \sim \mathcal{X}\times\mathcal{Y}} \big[ L \big(\tilde{y}(x|h),y\big)\big] \leq b
\end{split}
\end{equation}
We note that formulations based on Equation~\ref{eq:network_design} do not
incorporate design choices encountered in adaptive CNNs.

\subsection{Key insight behind our work}

In summary, the two aforementioned
optimization problems are decoupled and have not been jointly studied 
to date. On the one hand, the \emph{network selection} problem
(Equation~\ref{eq:problem_generic}) assumes the architectures
of CNNs (\emph{i.e.}, vectors $h$) to be fixed and that
the networks are already available and pretrained.
Consequently, the optimization problem in Equation~\ref{eq:problem_generic}
is solved only over possible $\kappa$'s. On the other hand, the \emph{network design}
problem (Equation~\ref{eq:network_design}) assumes a single
monolithic CNN and cannot be directly applied to
adaptive CNNs. In the following section, we describe
how our methodology elegantly formulates and optimizes
both problems jointly.

\section{Proposed Methodology}
\subsection{Hyper-parameter optimization}

We formulate the design of adaptive CNNs
as a hyper-parameter optimization problem with respect to both the
decision functions as well as the networks' sizing.
For the system of $M$ neural networks we can define
design space $\mathcal{Z}$ that consists of the set of hyper-parameters for each
network $h_i$ and hyper-parameters $\theta$'s
of the decision functions $\kappa$'s, \emph{i.e.},
$z = (h_1, ..., h_M, \theta_{12}, \theta_{23}, ...), z \in \mathcal{Z}$.

\textbf{Energy minimization}:
Our goal is to minimize the average (expected) energy
consumption per image subject to a maximum accuracy degradation.
Let us denote as $E(N_i(x|h_i))$ the energy consumed when executing neural
network $N_i$ to classify an input image $x \in \mathcal{X}$.
For the two-network system (\emph{i.e.}, $M=2$), we write:

\noindent
\begin{equation}
\label{eq:problem_en}
\begin{split}
\underset{z = (h_1, h_2, \theta_{12})}{\min}
\text{~~}  \mathbb{E}_{x \sim \mathcal{X}} \Big[ E_2 \big(N_2(x|h_2)\big) \cdot  
\kappa_{12}\big( & N_1(x|h_1) | \theta_{12} \big) + \\ & E_1 \big(N_1(x|h_1) \big)  \Big]\\
s.t. \text{~~} \mathbb{E}_{(x,y) \sim \mathcal{X}\times\mathcal{Y}}  \Big[
\Big( 1-\kappa_{12} \big(N_1(x|h_1) | & \theta_{12}\big) \Big) \cdot \\ 
\Big(L_1  \big( \tilde{y}_1(x|h_1),y \big) - L_2 & \big(\tilde{y}_2(x|h_2),y \big)  \Big) \Big] \leq B
\end{split}
\end{equation}

As a key contribution of our work, note
that in the proposed formulation (Equation~\ref{eq:problem_en})
the overall energy and accuracy explicitly depend on the
architecture of the networks based on the
values of the respective hyper-parameters $h_i$.
Moreover, please note that, unlike prior art that resorts in exhaustive (offline) or
iterative (online) methods to find a value for the
threshold~\cite{parkBigLittle, takhirov2016classifier}, a hyper-optimization
treatment allows us to directly incorporate the $\theta$ values
as hyper-parameters that will be co-optimized alongside the sizing of the networks.

\textbf{Considering different design paradigms}:
Existing methodologies from several groups consider cases
that either (\emph{i}) all networks execute locally on the same
hardware platform~\cite{parkBigLittle, takhirov2016classifier}, and
(\emph{ii}) some of the networks are executed on remote servers,
hence some of the images will be communicated
over the web~\cite{Leroux2017}. The formulation of the objective function in
Equation~\ref{eq:problem_en} allows us to consider energy consumption
under different design paradigms.

First, let us consider the case where all networks are executed locally.
We denote as $\tau(N_i(x))$ and $P(N_i(x))$ the runtime and power required
when evaluating a neural network $N_i$ an input image $x \in \mathcal{X}$.
The expected (per image) energy is therefore written as:
\begin{equation}
\begin{split}
\mathbb{E}_{x \sim \mathcal{X}} \Big[ P_2\big(N_2(x| & h_2)\big)  \cdot \tau\big(N_2(x|h_2)\big)
\cdot \kappa_{12}\big( N_1(x|h_1) | \theta_{12}  \big) \\ & + P_1\big(N_1(x|h_1)\big) \cdot \tau\big(N_1(x|h_1)\big) \Big]
\end{split}
\end{equation}
Second, we consider the case where some of the images $x \in \mathcal{X}$
are sent to the server over the network and the result is being communicated back.
In this case, the energy consumed is the product of the power
consumed from the client node $P(N_i(x))$ and total
runtime $\tau(N_i(x))$ that takes to send the image and receive the result.
For the case of the two-network system, this design will correspond to
having $N_1$ executing on the edge device and $N_2$ on the server machine.
Hence, the total energy consumption is:
\begin{equation}
\begin{split}
 &  \mathbb{E}_{x \sim \mathcal{X}}  \Big[P_{idle}\cdot \Big( \tau_{server} \big(N_2(x|h_2) \big) + \tau_{communication} \Big) \cdot \\
 & \kappa_{12}\big( N_1(x|h_1) | \theta_{12} \big)  + P_1\big(N_1(x|h_1)\big) \cdot \tau_{edge}\big(N_1(x|h_1)\big) \Big]
\end{split}
\end{equation}
where $\tau_{edge}$ and $\tau_{server}$ is the runtime when
executing on the edge device and the server, respectively, and
$P_{idle}$ is the idle power of the edge device while waiting for the result.

\textbf{Energy-constrained optimization}: We can also consider the problem of
minimizing classification error, while not exceeding a maximum
energy per image $E_{max}$. This design case is significant for image classification
under the energy budget imposed by mobile hardware engineers.
Formally, we can solve:

\noindent
\begin{equation}
\label{eq:problem_acc}
\begin{split}
\underset{z = (h_1, h_2, \theta_{12})}{\min}
\text{~~}  \mathbb{E}_{(x,y) \sim \mathcal{X}\times\mathcal{Y}} \Big[
\Big(1-\kappa_{12}(N_1(x|h_1) | & \theta_{12})\Big) \cdot \\ \Big( L_1\big(\tilde{y}_1(x|h_1),y\big) -
L_2\big( & \tilde{y}_2(x|h_2),y\big) \Big) \Big] \\
s.t. \text{~~}
\mathbb{E}_{x \sim \mathcal{X}} \big[ E_2\big(N_2(x|h_2)\big) \cdot 
\kappa_{12}\big(N_1(x|h_1) | \theta_{12}\big) & + \\  E_1\big( N_1(x| & h_1) \big) \big]  \leq  E_{max}
\end{split}
\end{equation}

\subsection{Adapting Bayesian optimization}

Solving Equations~\ref{eq:problem_en} and~\ref{eq:problem_acc}
gives rise to a daunting optimization problem, due to the fact
that the constraint and objective have terms that are costly
to evaluate. That is, to obtain the overall accuracy, all
networks need to be trained, hence each step of an optimization
algorithm can take hours to complete.

\begin{algorithm}[h!]
\caption{Bayesian optimization framework}\label{alg:opt}
\begin{algorithmic}[1]
\Require Obj. function $f(z)$, constraint $g(z)$,
constraint value $c$, design space $\mathcal{Z}$, Num. iterations $D$
\Ensure Optimizer $z^*$ of adaptive neural networks
\vspace{+5pt}
\For {$d = 1, 2, ..., D$}
    \State $\mathcal{M}  \leftarrow $ fit models on data so far $\mathcal{D}$ \label{line:model}
    \State $z_d \leftarrow \argmax_{z \in \mathcal{Z}} \alpha(z, \mathcal{M})$ ~~// \emph{acquisition function max.} \label{line:acq}
    \State // \emph{Training and profiling each network $N_i$}
    \For {$i = 1, 2, ..., M$} \label{line:net_start}
        \State $L_{i} (z_{i,d}) \leftarrow $ train network $N_i$ \label{line:train}
        \State // \emph{power, runtime measurements on device}
        \State $E_{i} (z_{i}^{d}) \leftarrow $energy of $N_i$  \label{line:power}
    \EndFor \label{line:net_end}
    \State // \emph{Evaluate accuracy and energy consumption}
    \State $u^{d}, v^{d} \leftarrow $ evaluate obj. $f(z^{d})$ and const. $g(z^{d})$   \label{line:obj}
    \State $\mathcal{D} = \mathcal{D} \cap \{z^d, u^d, v^d\} $
    \State // \emph{$\kappa$ function fine-tuning}
    \For {each decision function  $\kappa_{i, i+1}$}
        \For {$\theta_{i, i+1}' = 0.0, ..., 1.0$} \label{line:theta_start}
            \State $z^d \leftarrow (h^d_1, ..., h^d_M, ..., \theta_{ij}', ...) $
            \State $u', v' \leftarrow $ evaluate $f(z')$ and $g(z')$
            \State $\mathcal{D} = \mathcal{D} \cap \{z', u', v'\} $
        \EndFor \label{line:theta_end}
    \EndFor
\EndFor
\State \Return $z^* \leftarrow \argmax_{z} \{u^1, ..., u^{D} \}
\text{~} s.t. \text{~}  v^* \leq c$
\end{algorithmic}
\end{algorithm}

To enable energy-aware hyper-parameter optimization, we use
Bayesian optimization. The effectiveness of Bayesian optimization comes from
of the approximation the costly ``black-box'' functions with a surrogate model,
based on Gaussian processes~\cite{snoek2012practical}, which is cheaper to evaluate.
The design of adaptive CNNs
(Equations~\ref{eq:problem_en},~\ref{eq:problem_acc}) corresponds to
constrained minimizing function $f(z)$ over
design space $\mathcal{Z}$ and constraint $g(z)$, \emph{i.e.},
${\min}_{{z \in \mathcal{Z}}} ~f(z)$, subject to $g(z) \leq c$.
We adapt Bayesian optimization in this design space and we
summarize the methodology in Algorithm~\ref{alg:opt}.

In each step $d$, Bayesian optimization queries the objective function $f$
and the constraint function $g$ at a candidate
point $z_{d}$ and records
observations $ u, v \in \mathcal{R} $, respectively, 
acquiring a  tuple $\langle z^d, u^d, v^d \rangle$. In the context of a
system of adaptive neural networks, this corresponds
to profiling the energy, runtime, and power of each network $N_i$
(line~\ref{line:power}), to obtain the expected energy consumption
and accuracy on an image dataset $\mathcal{X}$ (line~\ref{line:obj}).

In general, Bayesian optimization consists of three
steps: \emph{first}, the probabilistic models $\mathcal{M}$
(for $f$ and $g$) are fitted based on the set of $d-1$ data points
$\mathcal{D} \in \mathcal{Z} = \langle z^l, u^l, v^l \rangle_{l=1}^{d-1}$
collected so far (line~\ref{line:model}); \emph{second}, the probabilistic models $\mathcal{M}$ are
used to compute a so-called acquisition function $\alpha (z, \mathcal{M})$,
which quantifies the expected improvement of the objective function
at arbitrary points $z \in \mathcal{Z}$, conditioned on
the observation history $\mathcal{D}$ ($\alpha (\cdot)$ in
line~\ref{line:acq}); \emph{last}, $f$ and $g$ are evaluated at point
$z_d$, which is the current feasible optimizer of the
acquisition function (line~\ref{line:acq}).

\textbf{Fine-tuning the functions $\kappa$}:
We make the observation that if we maintain
the sizing of the networks considered in each outer iteration of 
Algorithm~\ref{alg:opt}, we can fine-tune across the decision
functions $\kappa$. This step (lines~\ref{line:theta_start}
-\ref{line:theta_end}), which in the case of
threshold-based $\kappa$'s corresponds to sweeping across
$\theta$ values, has negligible complexity compared
to the overall optimization overhead and it
is equivalent to the threshold-based optimization employed
by prior art (without changing the architecture of the 
networks~\cite{parkBigLittle}).

The benefit of the $\kappa$-based fine-tuning is twofold. First,
in earlier stages of Bayesian optimization more data are being appended to the
observation history $\mathcal{D}$ which improves the convergence
of Bayesian optimization. Second, in the later states of Bayesian optimization,
the $\kappa$-based optimization serves as fine-tuning around the
near-optimal region. Effectively, our methodology combines the design space
\emph{exploration} properties inherent to Bayesian optimization-based methods,
and the \emph{exploitation} scope of optimizing only
over the $\kappa$ functions. As confirmed in our results,
Algorithm~\ref{alg:opt} improves upon the designs
considered during Bayesian optimization.

\section{Experimental Setup}

To enable energy-aware Bayesian optimization of adaptive CNNs,
we implement the key steps of Algorithm~\ref{alg:opt}
on top of the \texttt{Spearmint} tool~\cite{snoek2012practical}.
For all considered cases, we employ Bayesian optimization for 50
objective function evaluations.
As commercial embedded board we use NVIDIA Tegra TX1,
on which we deploy and profile the candidate networks $N_i$'s.
To measure the energy and power, and runtime values,
we use the power sensors available on the board. 
The use of Tegra (\emph{i.e.}, ARM-based architecture) limits our choices
of the Deep Learning packages to \texttt{Caffe}~\cite{jia2014caffe}.
Nevertheless, energy profiling has been successfully employed to other DL platforms
(as shown in~\cite{cai2017neuralpower}), hence
our findings can be extended to tools
such as TensorFlow. We train the candidate networks on a
server machine with an NVIDIA GTX 1070.

To enable a representative comparison with prior art, we consider a two-network
system with CaffeNet as $N_1$ and VGG-19 as $N_2$, 
as in ~\cite{parkBigLittle}.
For a three-network case, we consider LeNet, CaffeNet, and VGG-19 as
$N_1$, $N_2$, and $N_3$, respectively. Without loss of generality,
we employ hyper-parameter optimization and we learn the structure of the
CaffeNet network, without changing the LeNet or VGG-19 networks.
For the convolution layers we vary the number of feature maps (32-448) and
the kernel size (2-5), and for the fully connected layers the number of units (500-4000).\footnote{The ranges are selected around the original CaffeNet hyper-parameters~\cite{jia2014caffe}. Each network is trained using its original learning hyper-parameters (\emph{e.g.,} learning rate), but these can be also treated as variables to solve for using Algorithm~\ref{alg:opt}, as in~\cite{stamoulis2017hyperpower}.}
Due to resource limitations (Tegra storage, runtime requirements of
Bayesian optimization, \emph{etc.}) we currently assess the proposed methodology
on CIFAR-10. In the future, we plan to investigate the
transferability to larger datasets (\emph{e.g.}, ImageNet) 
and platforms beyond commercial GPUs (\emph{e.g.,} hardware accelerators \cite{isscc_2016_chen_eyeriss}). In addition, we plan to explore the 
use of predictive models that span different types of deep learning
frameworks and commercial GPUs~\cite{cai2017neuralpower}, and
which could be flexibly extended to account for process variations~\cite{stamoulis2016canwe, stamoulis2015efficient, chen2017profit} and
thermal effects~\cite{cai2016exploring}.

We consider the following \textbf{test cases} of
adaptive image classification: (\emph{i}) all CNNs execute locally on the
mobile system and we denote this case as \emph{local};
(\emph{ii}) only $N_1$, \emph{i.e.}, the less complex network is deployed
on the mobile system (edge node), while the more accurate networks execute on a server
(\emph{remote} execution). For every image item, the decision function selects
whether to use the local prediction or to communicate the image
to the server and receive the result of the more complex
network. We compute the communication time to transfer
\texttt{jpeg} images and to receive the classification results back
over two types of connectivity, via Ethernet and over WiFi, which we denote
as \emph{Ethernet} and \emph{wireless} respectively.
As an interesting direction in the context of IoT
applications, our method can be flexibly extended to profile and
design for other compression and communication protocols
(\emph{e.g.}, 4G modules, Bluetooth, \emph{etc.}), whose
integration we leave for future work.

\section{Experimental Results}
\begin{figure*}[ht!]
    \centering
    \begin{subfigure}[t]{0.49\textwidth}
        \centering
        \includegraphics[width=1.0\columnwidth]{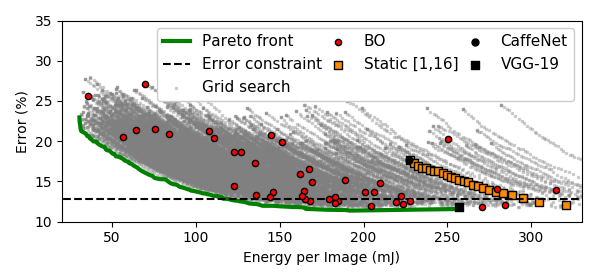}
                \vspace{-19pt}
          \caption{Embedded (local) execution for both networks.}
          \label{fig:design_space_evaluation_2_paper}
    \end{subfigure}%
    ~\hspace{10pt}
    \begin{subfigure}[t]{0.49\textwidth}
        \centering
        \includegraphics[width=1.0\columnwidth]{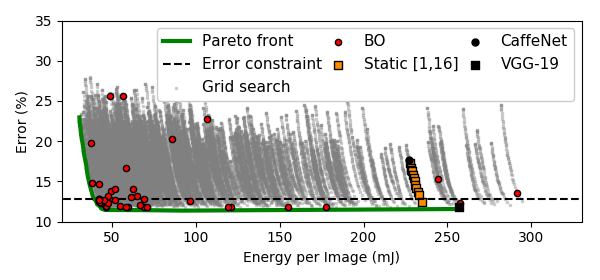}
        \vspace{-19pt}
          \caption{An edge-server energy-minimization design paradigm.}
          \label{fig:comm_design_space_evaluation_2_paper}
    \end{subfigure}
        \vspace{-7pt}
    \caption{Energy minimization under maximum error constraint.
          Bayesian optimization considers configurations (red circles) around the near-optimal region,
          while significantly outperforming static-design systems (orange squares).
          }
        \vspace{-2pt}
     \label{fig:design_space_evaluation_2_comparison_paper}
\end{figure*}

\textbf{Adaptive CNNs as hyper-parameter optimization problem}:
We first demonstrate the advantages that hyper-parameter optimization
offers in this design space. For both these cases of \emph{local} and
\emph{remote} execution with two networks, we employ grid search and we plot the obtained
error-energy pairs in Figures~\ref{fig:design_space_evaluation_2_paper}
and~\ref{fig:comm_design_space_evaluation_2_paper}, respectively.
We highlight this trade-off between accuracy and energy
by drawing the Pareto front (green line).\footnote{This visualization
is insightful since the constrained optimization problem under consideration
can be equivalently viewed as a multi-objective function, by writing the constraint term
as the Lagrangian (\emph{e.g.}, such formulation is
used in~\cite{bolukbasi17a}).}

\begin{figure*}[ht!]
    \centering
    \begin{subfigure}[t]{0.49\textwidth}
        \centering
          \includegraphics[width=1.0\columnwidth]{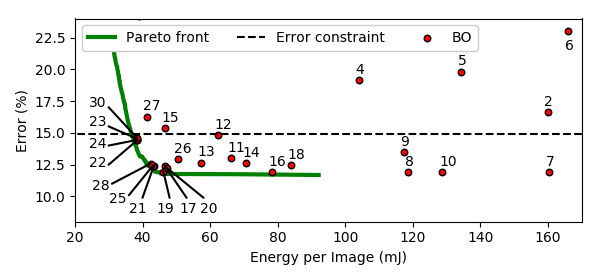}
        \vspace{-17pt}
            \caption{Sequence of configurations selection}
          \label{fig:comm_design_space_evaluation_BO_points}
    \end{subfigure}
    ~\hspace{10pt}
    \begin{subfigure}[t]{0.49\textwidth}
        \centering
        \includegraphics[width=1.0\columnwidth]{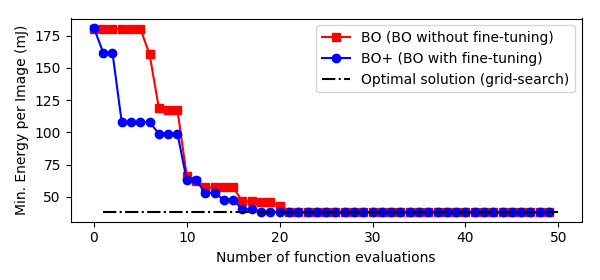}
        \vspace{-17pt}
        \caption{Best solution against the number of function evaluations.}
          \label{fig:comm_design_space_evaluation_BO_evolution}
    \end{subfigure}%
    \vspace{-5pt}
    \caption{Bayesian optimization for minimum energy under
          error constraints in the edge-server design. The method progressively
          evaluates designs closer to the Pareto front. The near-optimal region
          is reached within 22 function evaluations.}
    \label{fig:comm_design_space_evaluation_paper}
\end{figure*}

This illustration is insightful and allows us to make three important
observations. \emph{First}, we note how far to the left the Pareto
front is compared to the configurations obtained by existing works that
optimize only the decision function (orange squares). This fully captures
the motivation and novelty behind our work, \emph{i.e.}, to
treat the design of adaptive CNN systems as a hyper-parameter
problem and globally optimize for the architecture of the CNNs.

\emph{Second}, it is important to observe that the configurations considered
by Bayesian optimization (red circles) lie to the left of both the
static optimization designs (orange squares), as well as the monolithic
AlexNet and VGG-19 networks (black markers), hence showing that our methodology
allows for significant reduction in energy consumption (as discussed in detail next).
\emph{Third}, we note how close to the Pareto-front the Bayesian optimization points
are, showing the effectiveness of the method at eventually identifying
the feasible, near-optimal region.

\textbf{Evaluating the effectiveness of BO}: To fully assess
the effectiveness at reaching the near-optimal region in few tens of
function evaluations, we visualize the progress of BO for the case of
edge-server adaptive neural networks in Figure~\ref{fig:comm_design_space_evaluation_paper},
which corresponds to the edge-server design of Figure~\ref{fig:comm_design_space_evaluation_2_paper}.
First, we show the sequence of configurations selected in
Figure~\ref{fig:comm_design_space_evaluation_BO_points}, where we enumerate
the first 30 function evaluations. We observe that the near-optimal region
is reached with 22 Bayesian optimization steps.

Next, we assess the advantage that the $\kappa$-based fine-tuning
step offers. In Figure~\ref{fig:comm_design_space_evaluation_BO_evolution},
we show the minimum constraint-satisfying energy achieved during Bayesian
optimization without (red line) and with (blue line) this step employed
during the optimization. In the remainder of the results, we denote these
methods as BO and BO$^+$, respectively. We also show the optimal solution (dashed line),
as obtained by grid search. As motivated in our methodology section, we indeed
observe that the fine-tuning step allows the optimization to select
configurations around the near-optimal region in less function evaluations.
As we discuss in detail next, this is beneficial, especially in
over-constrained optimization cases.

\begin{figure*}[ht!]
    \centering
    \begin{subfigure}[t]{0.45\textwidth}
        \centering
        \hspace*{-22pt}
        \includegraphics[width=1.1\columnwidth]{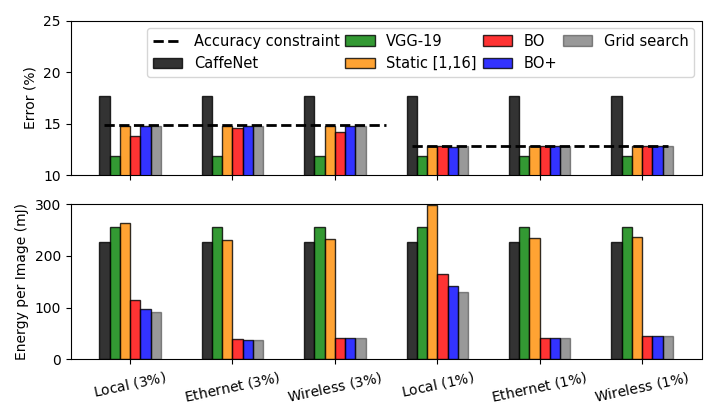}
          \vspace{-19pt}
          \caption{Energy minimization under maximum error constraints.}
          \label{fig:methods_comparison_everything_paper}
    \end{subfigure}%
    ~\hspace{10pt}
    \begin{subfigure}[t]{0.45\textwidth}
        \centering
        \includegraphics[width=1.1\columnwidth]{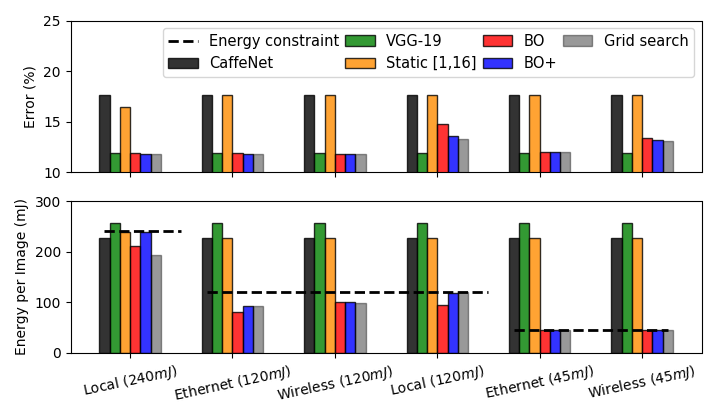}
        \vspace{-19pt}
          \caption{Error minimization under maximum energy constraints.}
          \label{fig:methods_comparison_everything_accuracy_paper}
    \end{subfigure}
        \vspace{-5pt}
    \caption{Assessing the effectiveness of the proposed methodology across different
    design paradigms and both constrained and over-constrained cases (the constraint
    values per case are given in the parentheses on the x-axis labels). We observe 
    that our methodology BO$^+$ (blue) successfully approaches the grid-search solution
    (gray), while always outperforming the best solution achieved by existing 
    static-design methods (orange).}
    \label{fig:methods_comparison_everything}
\end{figure*}

\textbf{Comprehensive adaptive CNNs evaluation}:
To enable a comprehensive analysis, we employ Bayesian optimization
on all three design practices, \emph{i.e.}, \emph{local},
\emph{Ethernet}, and~\emph{wireless}. For each design,
we solve both a constrained and an over-constrained case for
both the error-constrained energy minimization (Equation~\ref{eq:problem_en})
and the energy-constrained error minimization (Equation~\ref{eq:problem_acc})
problems. We plot the error on the validation set and energy
per-image in Figure~\ref{fig:methods_comparison_everything_paper}
and~\ref{fig:methods_comparison_everything_accuracy_paper},
respectively, for both BO and BO$^+$.
In addition, we compare the optimal result obtained by grid search, the best
result by previously published work that treats the CNNs
as blackboxes~\cite{parkBigLittle} (denoted as \emph{static}), and
the energy and error by solely using either $N_1$ (CaffeNet) or $N_2$ (VGG-19)
on their own. We report the constraints per case in the parentheses on the x axis.

\begin{table*}[ht!]
  \caption{Hyper-parameter optimization results on the test set.
  Our methodology does not overfit on the validation set.}
  \label{tab:hyperparams}
  \vspace{-7pt}
  \centering
  \scalebox{0.99}{
  \begin{tabular}{|c||c|c|c|c|c|c|}
    \hline
     ~  & \multicolumn{6}{|c|}{Minimum energy (mJ) achieved under allowed accuracy degradation constraint (given in parenthesis).}\\\hline
     Method   & Local ($3\%$) & Ethernet ($3\%$)  & Wireless ($3\%$) & Local ($1\%$)  & Ethernet ($1\%$) & Wireless ($1\%$) \\\hline
Static~\cite{bolukbasi17a, parkBigLittle} &  256.09 & 230.05 & 231.15 & 292.25 & 233.50 & 235.97  \\\hline
BO &  114.43 & 38.28 & 41.49 & 163.15 & 41.85 & 45.54  \\\hline
BO$^+$ &  99.35 & 38.00 & 40.50 & 148.26 & 41.68 & 45.54  \\\hline
Grid &  93.43 & 38.00 & 40.50 & 123.57 & 41.68 & 45.49  \\\hline\hline
     ~  & \multicolumn{6}{|c|}{Minimum error (\%) achieved under energy constraint (given in parenthesis).}\\\hline
     Method   & Local ($240mJ$) & Ethernet ($120mJ$)  & Wireless ($120mJ$) & Local ($120mJ$)  & Ethernet ($45mJ$) & Wireless ($45mJ$) \\\hline
Static~\cite{bolukbasi17a, parkBigLittle} &  17.30 & 18.44 & 18.44 & 18.44 & 18.44 & 18.44  \\\hline
BO &  12.67 & 12.77 & 12.74 & 20.69 & 12.83 & 14.42  \\\hline
BO$^+$ &  12.66 & 12.70 & 12.74 & 16.30 & 12.83 & 13.98  \\\hline
Grid &  12.58 & 12.58 & 12.58 & 14.14 & 12.82 & 13.98  \\\hline
  \end{tabular}
  }
\end{table*}

We note that the error percentage value corresponds to the maximum
accuracy degradation allowed compared to always using VGG-19
(\emph{i.e.}, the $B$ value in Equation~\ref{eq:problem_en}).
Last, in Figure~\ref{fig:methods_comparison_everything}
we report the validation error since this metric is the 
standard way of assessing the result of Bayesian optimization.
We also report the test error of the optimal, feasible 
solutions obtained in Table~\ref{tab:hyperparams}.
Based on the results, we can make several observations:

(\emph{i}) \textbf{Suboptimality of prior work}: As discussed
in the introduction, the energy consumption
of CaffeNet and VGG-19 shows a
small headroom for energy savings, thereby significantly limiting the effectiveness
of the static method. In fact, for \emph{local} energy minimization
with $1\%$ maximum accuracy degradation allowed (similar constraint as in~\cite{parkBigLittle}), statically designed adaptive CNNs will be forced to always use VGG-19 for the final prediction, 
while wastefully
evaluating CaffeNet first. This results in larger
energy consumption than using VGG-19 by itself,
as seen in Figure~\ref{fig:methods_comparison_everything_paper}.

(\emph{ii}) \textbf{Effectiveness proposed method}:
We observe that in all the considered cases the proposed BO$^+$ method
closely matches the result identified by grid search. 
For instance, in the over-constrained \emph{local} ($1\%$) 
case discussed before,
Bayesian optimization successfully considers a larger $N_1$ configuration
that trades-off energy consumption of $N_1$, yet improves the
overall accuracy such that the error constraint is satisfied.

In general, our methodology identifies designs that outperform
static methods~\cite{parkBigLittle} by up to $6 \times$ in terms
of minimum energy under accuracy constraints and
by up to $31.13\%$ in terms of error minimization under energy constraints.
Moreover, from Table~\ref{tab:hyperparams}, we confirm the generalization
of the methodology, \emph{i.e.}, that the adaptive
designs are near-optimal with respect to the test error and
that the method does not overfit on the validation set.

(\emph{iii}) \textbf{Enhanced performance via $\kappa$-based fine-tuning}:
As motivated in the methodology section, the fine-tuning step that we
incorporate in Bayesian optimization is beneficial especially in the
over-constrained cases, allowing the optimization to identify
near-optimal regions faster. In particular, BO$^+$ leads to
further energy minimization under accuracy constraints
by $13.18\%$ and to error minimization under energy constraints
by $21.22\%$, compared to the result of BO without fine-tuning.

(\emph{iv}) \textbf{Local versus remote execution}:
As an interesting finding in terms of design space exploration, we
observe that executing some of CNNs comprising the adaptive neural network
remotely allows for more energy-efficient image classification,
compared to executing everything locally at the same level
of accuracy. Such observations
have been also supported by recent design trends which ensure that,
due to privacy issues, mobile machine learning services maintain
both on-device and cloud components~\cite{Leroux2017, teerapittayanon2017distributed, ko2018edge}.

We observe this headroom from the Pareto front being further to the
left in Figure~\ref{fig:comm_design_space_evaluation_2_paper}
compared to Figure~\ref{fig:design_space_evaluation_2_paper}.
That is, using an edge-server design, where only the smallest CNN executes
locally, allows for energy reduction of $2.96 \times$
compared to executing all networks locally and for the same error constraint.
We postulate that this is an interesting finding in terms
of computation versus communication trade-offs and an interesting direction
to delve into.

Interestingly, we observe that the three-network case is
less energy-efficient than using two networks. This is to be
expected, since adaptive designs with more networks are
better fit for more computationally expensive applications, 
such as object recognition~\cite{bolukbasi17a}.
Exploring the sensitivity of performance metrics to the number of networks
is an interesting problem itself. Having shown that our approach can
reach near-optimal results compared to exhaustive grid search, we
leave this direction for future work.

\begin{table}[t!]
  \caption{Three-network case: Hyper-parameter optimization results on the test set.}
  \label{tab:hyperparams-3}
  \vspace{-10pt}
  \centering
  \scalebox{0.8}{
  \begin{tabular}{|c||c||c|}
    \hline
     ~  & Minimum energy (mJ) under & Minimum error (\%)\\
     Method  & accuracy degradation & under energy constraint: \\
     ~  & constraint: Local ($3\%$) & Local ($120mJ$)  \\\hline
Static~\cite{bolukbasi17a, parkBigLittle} &  268.43 & 41.25  \\\hline
BO &  137.36 & 17.94   \\\hline
BO$^+$ & 113.59 & 15.06  \\\hline
Grid &   108.45 & 14.98  \\\hline
  \end{tabular}
  }
\end{table}

\textbf{Exploring hierarchy}: Finally, we evaluate the
proposed method on a three-network case. We summarize the results
in Table~\ref{tab:hyperparams-3}. Once again, we observe that, compared to static
methods, our methodology closely matches the results obtained
by grid search. More specifically, in the case of energy minimization,
the solution reached by $BO^+$ is only $4.74\%$ away from the optimal
grid search-based design, while outperforming best previously
published static methods~\cite{parkBigLittle} by $2.47 \times$
in terms of energy minimization.

\section{Conclusion}

In this paper, we introduced an efficient hyper-parameter optimization
methodology to design hardware-constrained adaptive CNNs. The key
novelty in our work is that both the architecture settings of the CNNs
and the network selection problem are treated as
hyper-parameters to be globally (jointly) optimized.
To efficiently solve this problem,
we enhanced Bayesian optimization to the underlying properties of
the design space. Our methodology, denoted as $BO^+$, reached the
near-optimal region faster compared to a hardware-unaware BO,
and the optimal designs identified by grid search.

We exploited the effectiveness of the proposed methodology
to consider different adaptive CNN designs with respect to
energy, accuracy, and communication trade-offs and constraints
imposed by mobile devices.  Our methodology identified designs that
outperform previously published methods which use CNNs as
blackboxes~\cite{parkBigLittle} by up to $6 \times$ in terms of minimum energy per
image under accuracy constraints and by up to $31.13\%$ in terms
of error minimization under energy constraints.
Finally, we studied two image classification
practices, \emph{i.e.}, classifying all images locally versus
over the cloud under energy and communication constraints.

\begin{acks}
This research was supported in part by NSF CNS Grant No. 1564022 and by Pittsburgh Supercomputing Center via NSF CCR Grant No. 180004P.
\end{acks}

\bibliographystyle{ACM-Reference-Format}
\bibliography{dstam-iccad18}
%\bibliography{dstam-iccad18-short}

\end{document}